\documentclass[final]{cvpr}

\usepackage{times}
\usepackage{epsfig}
\usepackage{graphicx}
\usepackage{amsmath}
\usepackage{amssymb}
\usepackage{color}
\usepackage{algorithm}
\usepackage{algorithmic}
\usepackage{multirow}
\usepackage{booktabs}
\usepackage{pifont}
\usepackage{listings}
\usepackage{cite} 
\usepackage{soul}

\usepackage{tabulary,overpic,xcolor}
\usepackage[caption=false]{subfig}
\definecolor{citecolor}{RGB}{34,139,34}
\usepackage[pagebackref=true,breaklinks=true,letterpaper=true,colorlinks,
  citecolor=citecolor,bookmarks=false]{hyperref}

\usepackage[pagebackref=true,breaklinks=true,colorlinks,bookmarks=false]{hyperref}

\newcommand{\app}{\raise.17ex\hbox{$\scriptstyle\sim$}}

\newcolumntype{x}[1]{>{\centering\arraybackslash}p{#1pt}}

\newlength\savewidth\newcommand\shline{\noalign{\global\savewidth\arrayrulewidth
  \global\arrayrulewidth 1pt}\hline\noalign{\global\arrayrulewidth\savewidth}}
\newcommand{\tablestyle}[2]{\setlength{\tabcolsep}{#1}\renewcommand{\arraystretch}{#2}\centering\footnotesize}
\makeatletter\renewcommand\paragraph{\@startsection{paragraph}{4}{\z@}
  {.5em \@plus1ex \@minus.2ex}{-.5em}{\normalfont\normalsize\bfseries}}\makeatother

\def\cvprPaperID{5080} 
\def\confYear{CVPR 2021}

\begin{document}

\title{Exponential Moving Average Normalization \\
for Self-supervised and Semi-supervised Learning}

\author{Zhaowei Cai, Avinash Ravichandran, Subhransu Maji, Charless Fowlkes, Zhuowen Tu, Stefano Soatto\\
Amazon Web Services\\
{\tt\small \{zhaoweic,ravinash,smmaji,fowlkec,ztu,soattos\}@amazon.com}
}

\maketitle

\begin{abstract}
We present a plug-in replacement for batch normalization (BN) called exponential moving average normalization (EMAN), which improves the performance of existing student-teacher based self- and semi-supervised learning techniques. Unlike the standard BN, where the statistics are computed within each batch, EMAN, used in the teacher, updates its statistics by exponential moving average from the BN statistics of the student. This design reduces the intrinsic cross-sample dependency of BN and enhances the generalization of the teacher.  EMAN improves strong baselines for self-supervised learning by 4-6/1-2 points and semi-supervised learning by about 7/2 points, when 1\%/10\% supervised labels are available on ImageNet. These improvements are consistent across methods, network architectures, training duration, and datasets, demonstrating the general effectiveness of this technique. 
The code is available at https://github.com/amazon-research/exponential-moving-average-normalization.
\end{abstract}

\section{Introduction}
\label{sec:intro}

Supervised learning has achieved remarkable success on a variety of visual tasks,
benefiting from the availability of large-scale annotated datasets such as ImageNet \cite{DBLP:journals/ijcv/RussakovskyDSKS15}, MS-COCO \cite{DBLP:conf/eccv/LinMBHPRDZ14}, and ShapeNet \cite{chang2015shapenet}. However, in some domains such as medical imaging, large amounts of annotations are expensive or time-consuming to collect. Learning effective representations with small amounts (semi-supervised) or no (unsupervised or self-supervised) manual annotation is thus an important problem in computer vision \cite{lee2013pseudo,DBLP:conf/cvpr/WuXYL18,DBLP:conf/cvpr/He0WXG20,chen2020simple,chen2020big,grill2020bootstrap,DBLP:conf/iccv/BeyerZOK19,sohn2020fixmatch,DBLP:conf/nips/TarvainenV17,DBLP:conf/iclr/LaineA17}.

\begin{figure}[!t]
\centering
\centerline{\epsfig{figure=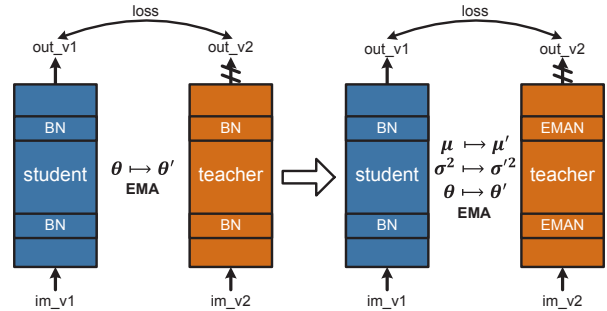,width=8cm,height=4.13cm}}
\caption{
The EMA-teacher framework with standard BN (left) and the proposed EMAN (right). $\theta$ are the model parameters, and $\mu$ and $\sigma^2$ BN statistics. EMA denotes exponential moving average updates. \texttt{im\_v1} and \texttt{im\_v2} are two different views of the same image. No gradient is backpropagated through the teacher model.
}
\label{fig:ema_teacher}\vspace{-3mm}
\end{figure}

Although many choices exist for semi- and self-supervised learning \cite{lee2013pseudo,DBLP:conf/eccv/NorooziF16,DBLP:conf/eccv/ZhangIE16,DBLP:conf/iclr/GidarisSK18,DBLP:conf/iccv/BeyerZOK19}, an effective approach is the family of student-teacher models \cite{DBLP:journals/corr/HintonVD15,chen2020big,DBLP:conf/nips/TarvainenV17,DBLP:conf/cvpr/He0WXG20,grill2020bootstrap,DBLP:conf/iclr/LaineA17,DBLP:conf/cvpr/XieLHL20}, where the outputs of the teacher are used to guide the learning of the student on the unlabeled data.
Within this family, a common approach is to update the teacher using exponential moving average (EMA) of the student parameters over its training trajectory \cite{DBLP:conf/nips/TarvainenV17}, which we call EMA-teacher, as shown in Figure \ref{fig:ema_teacher} (left).
As discussed in \cite{DBLP:conf/iclr/HuangLP0HW17,DBLP:conf/uai/IzmailovPGVW18,DBLP:conf/iclr/AthiwaratkunFIW19}, the temporally averaged teacher, as interpreted as the temporal ensembling of the student checkpoints, can improve generalization. Due to this property, it has been adopted in recent self-supervised learning methods \cite{DBLP:conf/cvpr/He0WXG20,grill2020bootstrap}.

While the objective and the update mechanisms are different for the student and the teacher, both networks use the standard batch normalization (BN) \cite{DBLP:conf/icml/IoffeS15}, as in the early EMA-teacher frameworks \cite{DBLP:conf/nips/TarvainenV17}.
However, this can lead to two potential problems:
\vspace{-2mm}
\begin{enumerate}
    \item {\em Cross-sample dependency}.
    This is an intrinsic property of BN where the output of a sample is dependent on all other samples in the same batch. This cross-sample information leakage may allow the model to ``cheat'' in semi- or self-supervised learning.
    To avoid this, some special designs on normalization were applied in \cite{henaff2019data,DBLP:conf/cvpr/He0WXG20,chen2020simple,grill2020bootstrap}. For example, \cite{henaff2019data} switched to layer normalization \cite{ba2016layer}; MoCo \cite{DBLP:conf/cvpr/He0WXG20} designed ShuffleBN where a mini-batch uses BN statistics from other randomly sampled mini-batch; and SimCLR \cite{chen2020simple} and BYOL \cite{grill2020bootstrap} used Synchronized BN (SyncBN).
    \vspace{-2mm}
    \item {\em Model parameter mismatch}. In the teacher network, its parameters are averaged from the student parameters of previous iterations, but the batch-wise BN statistics are instantly collected at current iteration.
    This could lead to potential mismatch between the model parameters and the BN statistics in the parameter space.
\end{enumerate}\vspace{-2mm}

We present a simple replacement for standard BN used in the EMA-teacher framework, called exponential moving average normalization (EMAN).
As shown in Figure \ref{fig:ema_teacher} (right), the EMAN statistics (mean $\mu'$ and variance $\sigma'^2$) in the teacher are exponentially moving averaged from the student BN statistics, similar to the other parameters.
The EMAN is simply a linear transform, without batch-wise statistics computation, and thus has removed cross-sample dependency presented in BN in the teacher.
Since the normalization statistics and model parameters are both updated using EMA, we expect this to improve stability of training by reducing the potential model parameter mismatches when using BN.
This simple design requires only a few lines of code, and can replace other complex normalization schemes (e.g. ShuffleBN, SyncBN, etc.) within various semi- and self-supervised learning techniques.

We have evaluated EMAN within various EMA-teacher frameworks, including recent state-of-the-art semi-supervised learning (FixMatch \cite{sohn2020fixmatch}) and self-supervised learning (MoCo \cite{DBLP:conf/cvpr/He0WXG20} and BYOL \cite{grill2020bootstrap}) techniques.
On self-supervised learning, EMAN improves the performance of MoCo/BYOL by 4-6/1-2
points when 1\%/10\% labels are available on ImageNet \cite{DBLP:journals/ijcv/RussakovskyDSKS15}.
On semi-supervised learning, EMAN improves the performance of FixMatch by about 7/2 points for 1\%/10\% labels, leading to the new state-of-the-art performances of 63.0/74.0 top-1 accuracy for 1\%/10\% labels on ImageNet.
These improvements are consistent across methods, network architectures, training duration, and datasets, demonstrating the effectiveness of EMAN as a general technique.
In addition, EMAN is just as efficient as standard BN, and does not require cross-GPU communication or synchronization of ShuffleBN or SyncBN.
We thus believe that EMAN can be of interest for other future student-teacher variants.

\section{Related Work}

\textbf{Semi-supervised learning}  leverages unlabeled data to improve the model performance, and has a long history in machine learning \cite{zhu2005semi,chapelle2009semi}. We primarily focus on recent deep-learning based approaches.
Pseudo-Labeling \cite{lee2013pseudo} generates synthetic labels from the confident predictions to learn on the unlabeled data.
Temporal ensembling of predictions was proposed to improve robustness in~\cite{DBLP:conf/iclr/LaineA17}.
Consistency regularization based methods~\cite{DBLP:conf/iclr/LaineA17,DBLP:conf/nips/TarvainenV17,sohn2020fixmatch,DBLP:journals/pami/MiyatoMKI19} learn by requiring the predictions to be consistent after perturbations on inputs and/or model parameters. For example, $\Pi$-model \cite{DBLP:conf/iclr/LaineA17} perturbs the model weights, uses dropout \cite{DBLP:journals/jmlr/SrivastavaHKSS14}, and enforces that the clean and noisy predictions be consistent.
Mean-teacher \cite{DBLP:conf/nips/TarvainenV17} proposed the EMA-teacher framework, and learns by enforcing consistency between the student and teacher models.
FixMatch \cite{sohn2020fixmatch} assumes consistency between the weakly and strongly augmented inputs.
A broader survey of semi-supervised learning techniques can be found in~\cite{zhu2005semi,chapelle2009semi}.

\textbf{Unsupervised or self-supervised learning} aims to learn representations from data without annotations.
It has been particularly effective in natural language processing \cite{DBLP:conf/naacl/DevlinCLT19,radford2018improving}.
Early self-supervised learning approaches in computer vision were based on proxy tasks, e.g. solving jigsaw puzzles \cite{DBLP:conf/eccv/NorooziF16}, colorization \cite{DBLP:conf/eccv/ZhangIE16} and rotation prediction \cite{DBLP:conf/iclr/GidarisSK18}.
Recently, the contrastive learning\cite{DBLP:conf/cvpr/HadsellCL06} using instance discrimination has achieved promising results~\cite{DBLP:conf/cvpr/WuXYL18,tian2019contrastive,DBLP:conf/cvpr/He0WXG20,DBLP:conf/cvpr/MisraM20,chen2020simple,chen2020big}.
For example, MoCo \cite{DBLP:conf/cvpr/He0WXG20} and SimCLR \cite{chen2020simple,chen2020big} have narrowed the gap between supervised and unsupervised learning in some domains.
BYOL \cite{grill2020bootstrap} found that, instead of a contrastive loss, optimizing a feature regression loss can achieve better results than prior work\cite{DBLP:conf/cvpr/He0WXG20,chen2020simple,chen2020big}. An extensive survey of self-supervised learning can be found in \cite{jing2020self}.

The \textbf{student-teacher framework} was first introduced in \cite{bucilu2006model} and developed in \cite{DBLP:journals/corr/HintonVD15} to distill knowledge from the pretrained teacher model to the new student model.
While in \cite{DBLP:journals/corr/HintonVD15,chen2020big}, the teacher is a pretrained and frozen model, other variants are available for different purposes. For example, in \cite{sohn2020fixmatch} the teacher and the student are identical; in \cite{DBLP:conf/aaai/ShenHX19} the teacher is an ensemble of multiple networks; in \cite{bucilu2006model,DBLP:journals/corr/HintonVD15,chen2020big} the teacher is a more complex network than the student for model compression; in \cite{DBLP:conf/iclr/LaineA17} the teacher is a temporal ensemble of student checkpoints with the step of one epoch; in \cite{DBLP:conf/nips/TarvainenV17,DBLP:conf/cvpr/He0WXG20,grill2020bootstrap}, the teacher is a more smoothly temporal ensemble than \cite{DBLP:conf/iclr/LaineA17} by exponential moving average.

\textbf{Normalization} is a
critical component to enable faster convergence and reduce the dependency on initialization for modern deep networks. While BN \cite{DBLP:conf/icml/IoffeS15} is widely used, it introduces some issues, such as requiring large batch sizes for accurate statistics, and mismatch between how BN is used during training and inference.
To address these, other normalization techniques have been proposed.
Layer Normalization (LN) \cite{ba2016layer} normalizes along the channel and spatial dimension, Instance Normalization (IN) \cite{ulyanov2016instance} along only the spatial dimension, and Group Normalization (GN) \cite{DBLP:conf/eccv/WuH18} operates similar to LN but divides the channels into groups.
MABN \cite{Yan2020Towards} shares some similarities with our EMAN, but mainly focuses on the stability of small batch size training and updates its statistics inside a single network.
In self-supervised learning, to avoid the possible information leakage via BN, \cite{henaff2019data} used LN, SimCLR \cite{chen2020simple} and BYOL \cite{grill2020bootstrap} use SyncBN, and MoCo \cite{DBLP:conf/cvpr/He0WXG20} uses ShuffleBN where a mini-batch uses BN statistics from other randomly sampled mini-batch.
Although these normalization schemes work well in some specific cases, our experiments will show that they do not generalize well across various semi- and self-supervised learning methods.

\section{Preliminaries}

\subsection{EMA-Teacher Framework}
\label{subsec:ema-teacher}

The EMA-teacher framework, with architecture shown in Figure \ref{fig:ema_teacher} (left), was first introduced in the Mean Teacher \cite{DBLP:conf/nips/TarvainenV17}, to improve the non-smooth temporal ensembling of \cite{DBLP:conf/iclr/LaineA17}. The teacher parameters $\theta'$ are updated by exponential moving average (EMA) from the student parameters $\theta$,
\begin{equation}
\label{equ:ema update}
    \theta':=m\theta'+(1-m)\theta,
\end{equation}
where the momentum $m$ is a number close to 1, e.g. 0.999. The student network is exactly the same as the standard supervised network, where the parameters $\theta$ are learned by standard SGD. In general, there is no gradient backpropagation through the teacher model, and the teacher model is discarded once training finished.

This EMA teacher can be interpreted as a smooth temporal ensembling of the student checkpoints along the training trajectories. As discussed in \cite{DBLP:conf/iclr/HuangLP0HW17,DBLP:conf/uai/IzmailovPGVW18,DBLP:conf/iclr/AthiwaratkunFIW19}, this temporal weight averaging mechanism can stabilize training trajectories and present better performances than the standard SGD update. In consistency based semi- and self-supervised learning, training could be less stable \cite{DBLP:conf/iclr/AthiwaratkunFIW19}, where the EMA-teacher framework with improved generalization can help. Due to its good performance, this EMA-teacher has derived different variants for different tasks \cite{DBLP:conf/cvpr/He0WXG20,grill2020bootstrap}.

While the EMA-teacher has the special update rule for the learnable parameters, it does not for its normalization operators. Instead, the standard BN is used in both student and teacher models as in \cite{DBLP:conf/nips/TarvainenV17}.

\subsection{Batch Normalization}

BN \cite{DBLP:conf/icml/IoffeS15} can stabilize the learning and enable faster convergence, and thus has been widely adopted. It has different training and inference modes. During training, BN first computes the mean and the variance of the layer inputs for the current batch $\{x_i\}_{i=1}^{n}$,
\begin{equation}
\label{equ:bn stat}
\begin{aligned}
    \mu_\mathcal{B}&=\frac{1}{n}\sum_{i=1}^{n}x_i, \\
    \sigma_\mathcal{B}^2&=\frac{1}{n}\sum_{i=1}^{n}(x_i-\mu_\mathcal{B})^2,
\end{aligned}
\end{equation}
where $n$ is batch size. Next, every sample $x$ in the current batch is normalized using the batch-wise statistics $\mu_\mathcal{B}$ and $\sigma_\mathcal{B}^2$, and then an affine transformation with learnable parameters $\gamma$ and $\beta$ is applied,
\begin{equation}
\label{equ:bn forward}
    \hat{x}=BN(x)=\gamma\frac{x-\mu_\mathcal{B}}{\sqrt{\sigma_\mathcal{B}^2+\epsilon}}+\beta,
\end{equation}
where $\epsilon$ is a small constant for numerical stability.

At inference, however, it is not desirable to use the batch-wise statistics, $\mu_\mathcal{B}$ and $\sigma_\mathcal{B}^2$, since the output of an input should be deterministic and not dependent on other inputs in the same batch. The population statistics, $E[\mu]$ and $E[\sigma^2]$, should be used instead. But this requires an additional stage of statistics gathering on a large sample population, which could be undesirable. In many implementations, a more practical and efficient strategy is widely used, collecting the proxy statistics $\mu$ and $\sigma^2$ by exponential moving average during training,
\begin{equation}
\label{equ:momentum stat update}
\begin{aligned}
    \mu&:=\alpha\mu+(1-\alpha)\mu_\mathcal{B}, \\
    \sigma^2&:=\alpha\sigma^2+(1-\alpha)\sigma_\mathcal{B}^2,
\end{aligned}
\end{equation}
where the momentum $\alpha$ here is usually 0.9. With the proxy statistics $\mu$ and $\sigma^2$, the BN at inference becomes
\begin{equation}
\label{equ:BN inference}
    \hat{x}=BN(x)=\gamma\frac{x-\mu}{\sqrt{\sigma^2+\epsilon}}+\beta,
\end{equation}
which differs from its training mode of (\ref{equ:bn forward}). This practical strategy is very common in many implementations, e.g. as default in PyTorch and TensorFlow.

\section{Exponential Moving Average Normalization}
\label{sec:eman}

In the EMA-teacher framework, as introduced in Section \ref{subsec:ema-teacher}, both the student and the teacher use the standard BN during training,
\begin{equation}
\begin{aligned}
    y&=f(BN(x),\theta), \\
    y'&=f(BN(x),\theta').
\end{aligned}
\end{equation}
where $f$ is the intermediate layers of \texttt{relu-conv}, which takes the output of normalization as input.
The standard BN is well aligned with the model parameters for a typical network (e.g. the student) which is updated by SGD, since the parameters are optimized with those batch-wise statistics. However, it is no longer the case for the teacher that is updated by EMA. Two reasons suggested that. First, the teacher is used to generate pseudo ground-truth to guide the learning of the student. With batch-wise BN, these generated pseudo labels will be cross-sample dependent, which is not desirable. For example, the pseudo label of $x_1$ is dependent on $x_2$ if $x_1$ and $x_2$ are in the same training batch. Second, there is a possible mismatch between the model parameters $\theta'$ and batch-wise BN statistics ($\mu_\mathcal{B}$ and $\sigma_\mathcal{B}^2$) in the teacher model. The former is averaged from the student parameters of previous iterations, but the latter is instantly collected at current iteration, and the former is not optimized for the latter. This mismatch could lead to non-smoothness in the parameter space.

To resolve these issues, we propose using exponential moving average normalization (EMAN) for the teacher during training (student still uses BN),
\begin{equation}
    y'=f(EMAN(x),\theta'),
\end{equation}
where
\begin{equation}
\label{equ:eman forward}
    \hat{x}=EMAN(x)=\gamma\frac{x-\mu'}{\sqrt{\sigma'^2+\epsilon}}+\beta,
\end{equation}
where $\mu'$ and $\sigma'^2$ are also exponentially moving averaged from the student $\mu$ and $\sigma^2$, in the same way of (\ref{equ:ema update}),
\begin{equation}
\begin{aligned}
\label{equ:eman update}
    \mu'&:=m\mu'+(1-m)\mu, \\
    \sigma'^2&:=m\sigma'^2+(1-m)\sigma^2.
\end{aligned}
\end{equation}
The key difference between (\ref{equ:bn forward}) and (\ref{equ:eman forward}) is the normalization factors. They are batch-wise $\mu_\mathcal{B}$ and $\sigma_\mathcal{B}^2$ in (\ref{equ:bn forward}), but EMA updated $\mu'$ and $\sigma'^2$ in (\ref{equ:eman forward}). This new normalization technique for the teacher is simply a linear transform which is no longer dependent on batch statistics. EMAN eliminates cross-sample dependence in the teacher, and there is no mismatch between the model parameters ($\theta'$) and its normalization factors ($\mu'$ and $\sigma'^2$).
Note that although the student is still cross-sample dependent, this is a less serious issue than the cross-sample dependency in the teacher.
EMAN is better aligned with the EMA-teacher framework than the standard BN (and probably other normalization), and as we show next, it is generally applicable in different EMA-teacher variants for different tasks \cite{DBLP:conf/nips/TarvainenV17,sohn2020fixmatch,DBLP:conf/cvpr/He0WXG20,grill2020bootstrap}.

\begin{algorithm}[t]
\caption{PyTorch-like Pseudocode of EMAN Update}
\label{alg:eman}\vspace{-1mm}
\definecolor{codeblue}{rgb}{0.25,0.5,0.5}
\lstset{
  backgroundcolor=\color{white},
  basicstyle=\fontsize{7.2pt}{7.2pt}\ttfamily\selectfont,
  columns=fullflexible,
  breaklines=true,
  captionpos=b,
  commentstyle=\fontsize{7.2pt}{7.2pt}\color{codeblue},
  keywordstyle=\fontsize{7.2pt}{7.2pt},
}
\begin{lstlisting}[language=python]
# f_s, f_t: encoder networks for student and teacher

params_s = f_s.parameters() # learnable parameters
params_t = f_t.parameters() # learnable parameters
for s, t in zip(params_s, params_t):
    t = momentum*t + (1-momentum)*s

buffers_s = f_s.buffers() # BatchNorm proxy statistics
buffers_t = f_t.buffers() # BatchNorm proxy statistics
for s, t in zip(buffers_s, buffers_t):
    t = momentum*t + (1-momentum)*s
\end{lstlisting}\vspace{-1mm}
\end{algorithm}

\subsection{Applications}

We have applied EMAN to recent state-of-the-art semi-supervised learning (FixMatch \cite{sohn2020fixmatch}) and self-supervised learning (MoCo \cite{DBLP:conf/cvpr/He0WXG20} and BYOL \cite{grill2020bootstrap}) methods. Applying EMAN to these three techniques is simple, requiring a few lines of code change, as shown in Algorithm \ref{alg:eman}, where the learnable parameter update is adopted as in \cite{DBLP:conf/nips/TarvainenV17,DBLP:conf/cvpr/He0WXG20,grill2020bootstrap}.

\begin{figure}[!t]
\centering
\centerline{\epsfig{figure=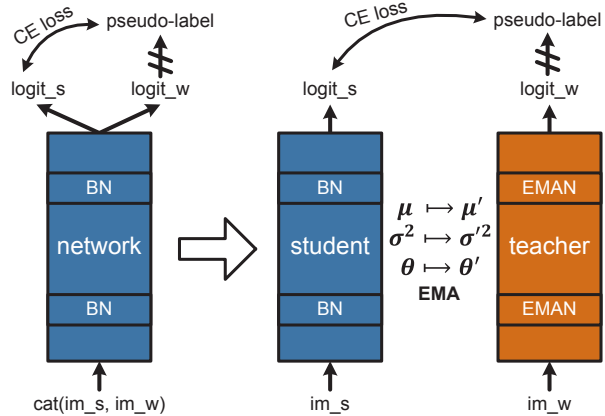,width=8cm,height=5.55cm}}
\caption{The architecture change of FixMatch using EMAN.
\texttt{im\_s}/\texttt{im\_w} is the strongly/weakly augmented view of an image, \texttt{cat} concatenation. The other symbols are similar as Figure \ref{fig:ema_teacher}.
}
\label{fig:fixmatch arch}\vspace{-3mm}
\end{figure}

\textbf{FixMatch} \cite{sohn2020fixmatch} uses identical teacher and student models, with architecture shown in Figure \ref{fig:fixmatch arch} (left). The teacher generates pseudo labels after thresholding, which are then used to guide the learning of the student with standard cross-entropy loss. A tricky mechanism in FixMatch is to concatenate the strongly and weakly augmented images first and then forward them to the model together. In this case, the teacher and the student are using exactly the same BN statistics. We first reframe FixMatch in the EMA-teacher framework (with standard BN), motivated by its success. However, this change leads to much worse performance, with possible reasons discussed above in this section.

\begin{figure*}[t]
\begin{minipage}[b]{.33\linewidth}
\centering
\centerline{\epsfig{figure=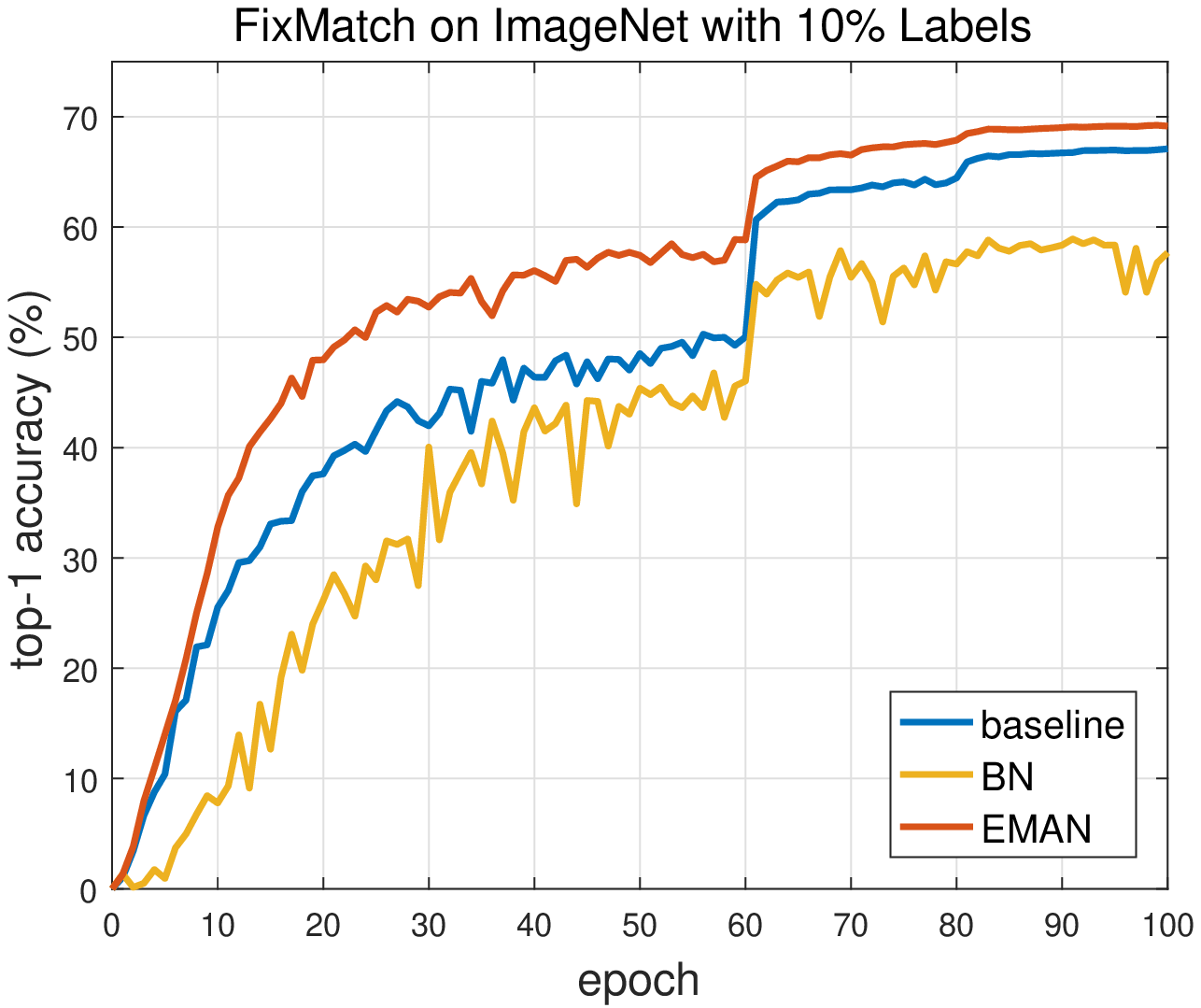,width=6cm,height=4.5cm}}{(a)}
\end{minipage}
\hfill
\begin{minipage}[b]{.33\linewidth}
\centering
\centerline{\epsfig{figure=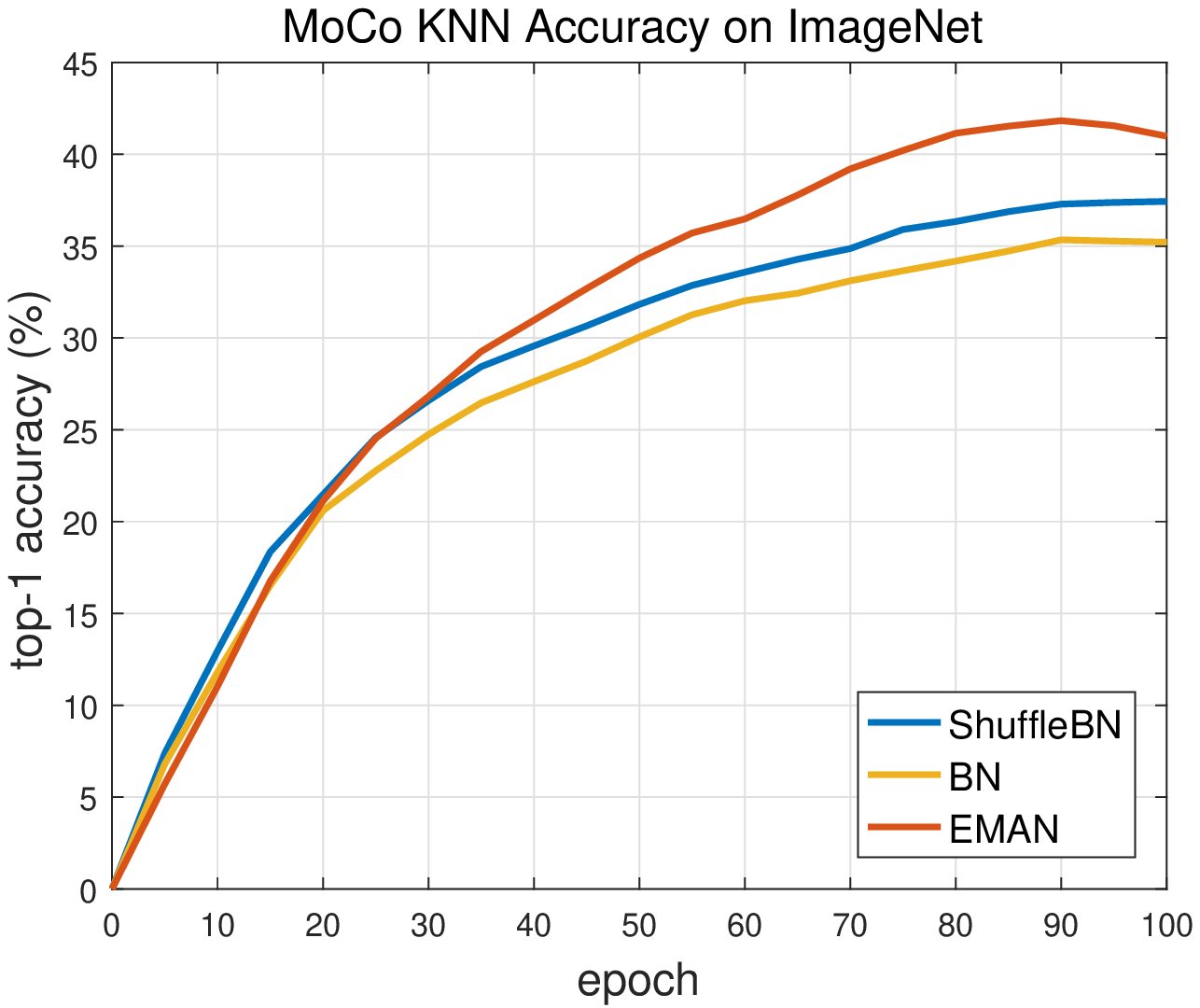,width=6cm,height=4.5cm}}{(b)}
\end{minipage}
\hfill
\begin{minipage}[b]{.33\linewidth}
\centering
\centerline{\epsfig{figure=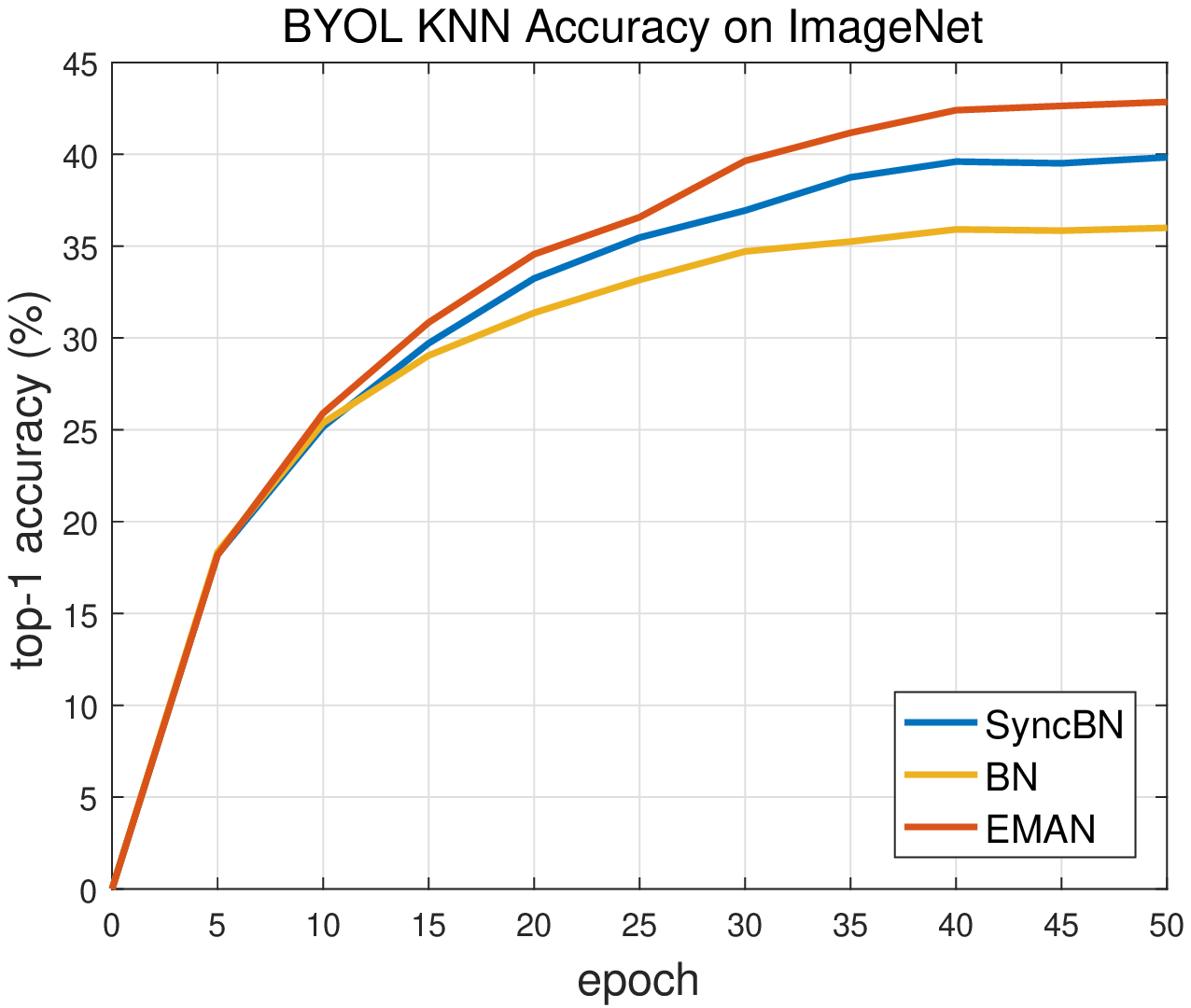,width=6cm,height=4.5cm}}{(c)}
\end{minipage}
\caption{The training accuracy curves of FixMatch, MoCo and BYOL on ImageNet, by using different normalization schemes.
}
\label{fig:ablation}\vspace{-1mm}
\end{figure*}

\textbf{MoCo} \cite{DBLP:conf/cvpr/He0WXG20} has bridged the gap between supervised and unsupervised learning in multiple visual tasks. It can be interpreted as a variant of EMA-teacher,
where the key (teacher) model is EMA updated from the query (student) model, and a contrastive loss is constructed between their outputs.
MoCo also found it problematic to use BN in both student and teacher, due to possible information leakage.
The model would probably ``cheat'' with local BN statistics to find a low-loss trivial solution rather than learning good representations.
Instead, MoCo uses ShuffleBN in the teacher, in which the batch-wise BN statistics are computed inside a randomly shuffled mini-batch samples across distributed GPUs. This ensures that the batch statistics used to compute the query and the key come from two different subsets, avoiding the cheating issue to some extent.

\textbf{BYOL} \cite{grill2020bootstrap} can also be interpreted as a EMA-teacher variant similar to MoCo, although the student/teacher is named as online/target network.
It differs from the other contrast based self-supervised learning \cite{DBLP:conf/cvpr/WuXYL18,DBLP:conf/cvpr/He0WXG20,chen2020simple} by formulating the self-supervised learning problem as a regression task, bridging the student and teacher outputs with a simple L2 loss. \cite{UnderstandingBYOL} hypothesizes that the reason why BYOL does not need contrastive loss is BN also plays an role of implicit contrast term, not just normalization. To have stronger implicit contrast and avoid knowledge leakage, SyncBN is adopted in both student and teacher models, in which the BN statistics are collected globally across GPU cards and machines. This requires efficient synchronization technique and leads to slower training speed.

Our experiments show that using standard BN in both teacher and student models results in poor performances in all these three techniques.
Although different solutions have been proposed to avoid that, e.g. Shuffle BN in MoCo and SycnBN in BYOL, they do not generalize well in other techniques as will be shown in our experiments. To have a general and simpler solution, we apply EMAN in all three techniques, as in Figure \ref{fig:ema_teacher} and \ref{fig:fixmatch arch}.
EMAN can improve over the standard BN by a large margin, and even surpass the ShuffleBN/SyncBN counterparts, universally in FixMatch/MoCo/BYOL. In addition, the training will be simpler and more efficient
since EMAN requires no cross-GPU communication or synchronization as needed in ShuffleBN/SyncBN.
We expect EMAN to be applicable to other student-teacher variants.

\section{Experiments}

ImageNet \cite{DBLP:journals/ijcv/RussakovskyDSKS15} is mainly used in all experiments, which contains $\sim$1.28 million images for training and 50K images for validation. The proposed EMAN has been evaluated on the state-of-the-art self-supervised learning (MoCo \cite{DBLP:conf/cvpr/He0WXG20} and BYOL \cite{grill2020bootstrap}) and semi-supervised learning (FixMatch \cite{sohn2020fixmatch}). For MoCo, the official implementation was used, but FixMatch and BYOL were reimplemented in PyTorch \cite{DBLP:conf/nips/PaszkeGMLBCKLGA19}. The default network is ResNet-50 \cite{DBLP:conf/cvpr/HeZRS16} and the default hyperparameters in the corresponding papers were used, unless  noted otherwise. For FixMatch, the batch size for labeled (unlabeled) images is 64 (320) with initial learning rate 0.03. For BYOL, the batch size is 512 with initial learning rate 0.9. All experiments were run on a machine with 8 V100 GPU cards. The self-supervised pretrained models were evaluated by 1) linear classification following \cite{DBLP:conf/cvpr/He0WXG20,chen2020simple,chen2020big,grill2020bootstrap}; and 2) kNN classification with $k=20$ following \cite{DBLP:conf/cvpr/WuXYL18,DBLP:conf/iccv/ZhuangZY19,li2020prototypical,caron2020unsupervised}, on top of the frozen representation. The other settings will be introduced in the following specific experimental sections. More experimental details can be found in the appendix.

\subsection{The Effect of EMAN}
\label{subsec:eman effect}

The effect of the proposed EMAN was evaluated. For FixMatch, only 10\% labels were used and the rest data as unlabeled. For MoCo and BYOL, we showed the accuracies of the kNN classifier along the training, since it is too expensive to train additional linear classifier. The kNN classifier used 10\% \texttt{train} (50\% \texttt{val}) as training set (query) for efficiency purpose (the observations are consistent with all \texttt{train}/\texttt{val} data). FixMatch/MoCo/BYOL was trained for 100/100/50 epochs, where FixMatch drops learning rate by 10 times at 60th and 80th epoch, and MoCo/BYOL uses cosine learning schedule. All training uses linear warm-up learning rate for 5 epochs.

\begin{table}[t]
\tablestyle{1.8pt}{1.2}
\begin{tabular}{x{42}x{42}|x{34}x{32}x{32}}\shline
student &teacher &FixMatch &MoCo &BYOL\\[.1em]\shline
default &default &67.1 &54.4  &55.4\\
BN &BN &58.9 &52.5  &52.0\\
SyncBN &SyncBN &52.0 &53.3  &55.4\\
BN &ShuffleBN &55.8 &54.4  &52.6\\
GN &GN &63.3 &49.3  &failed\\
IN &IN &61.3 &46.5  &failed\\
BN &EMAN &\bf{69.2} &\bf{55.8}  &\bf{56.2}\\[.1em]\shline
\end{tabular}\vspace{2mm}
\caption{Accuracy with different normalization.}
\label{tab:norm}\vspace{-3mm}
\end{table}

FixMatch was reframed to the EMA-teacher framework as in Figure \ref{fig:fixmatch arch}, using standard BN, denoted as ``BN'' in Figure \ref{fig:ablation} (a). However, this architecture change leads to much worse performance than the baseline FixMatch (``baseline''). Switching to standard BN also leads to worse performance than the baseline MoCo (ShuffleBN) and BYOL (SyncBN), as shown in Figure \ref{fig:ablation} (b) and (c). By simply changing the standard BN to the proposed EMAN in the teacher model, significant boosts are available in all FixMatch/MoCo/BYOL, e.g. roughly 10/6/7 points. This simple change also surpassed all three very strong baseline FixMatch/MoCo/BYOL by about 2/4/3 points.

To check the generalization, SyncBN and ShuffleBN were also evaluated in the other techniques, as shown in Table \ref{tab:norm}, where MoCo and BYOL were measured by linear classification on 10\% labeled data. Although they work well within their own technique (i.e., ShuffleBN in MoCo and SyncBN in BYOL), they do not generalize very well across techniques. For example, SyncBN is 1.1 points worse than ShuffleBN in MoCo and even 6.9 points worse than BN in FixMatch; and ShuffleBN is 2.8 points worse than SyncBN in BYOL and even 3.1 points worse than BN in FixMatch. In contrast, EMAN generalizes very well in all three techniques and achieved the best results. Other cross-sample independent normalization techniques were also evaluated in Table \ref{tab:norm}, including Group Normalization (GN) \cite{DBLP:conf/eccv/WuH18} and Instance Normalization (IN) \cite{ulyanov2016instance}. But they all lead to inferior performances. Also note that EMAN is as simple as BN, unlike ShuffleBN in MoCo and SyncBN in BYOL which rely on cross-GPU communication or synchronization. For example, switching SyncBN to EMAN in BYOL, the training can be speeded up by about 30\% with PyTorch implementation on a machine with 8 GPUs.

\begin{table}[t]
\tablestyle{1.8pt}{1.2}
\begin{tabular}{llx{22}x{22}x{22}x{22}x{22}x{22}}\shline
\multicolumn{2}{l}{\multirow{2}{*}{Method}} &\multicolumn{2}{c}{1\% labels} &\multicolumn{2}{c}{10\% labels} &\multicolumn{2}{c}{100\% labels}\\
&& top-1 & top-5 & top-1 &top-5 &top-1 &top-5\\[.1em]\shline
\multicolumn{2}{l}{Supervised \cite{DBLP:conf/cvpr/HeZRS16,DBLP:conf/iccv/BeyerZOK19}} &25.4 &48.4  &56.4 &80.4 &76.1  &92.9\\ [.1em]\shline
\parbox[t]{2mm}{\multirow{8}{*}{\rotatebox[origin=c]{90}{Linear}}} &MoCo &43.2 &71.0  &58.8 &82.6 &67.5  &88.1\\
&MoCo-EMAN &48.9 &75.3  &60.5 &83.5 &67.7  &88.0\\\cline{2-8}
&MoCo (2$\times$) &51.5 &77.6  &64.2 &86.0 &72.4  &90.9\\
&MoCo-EMAN (2$\times$) &56.8 &80.4  &65.7 &86.4 &72.3  &90.6\\\cline{2-8}
&MoCo (800) &50.4  &76.6  &63.0  &85.4 &70.3  &90.0\\
&MoCo-EMAN (800) &55.4  &79.3  &64.0  &85.3  &70.1  &89.3\\\cline{2-8}
&BYOL &51.3 &76.3  &64.8 &86.2 &71.4  &90.2\\
&BYOL-EMAN &55.1 &78.9  &66.7 &87.3 &72.2  &90.7\\ [.1em]
\shline
\parbox[t]{2mm}{\multirow{8}{*}{\rotatebox[origin=c]{90}{Finetune}}} &MoCo &44.8 &73.4  &63.3 &86.1 &76.1  &92.9\\
&MoCo-EMAN &50.4 &77.8  &64.9 &87.1 &76.0  &93.0\\\cline{2-8}
&MoCo (2$\times$) &53.1  &79.9  &67.9  &88.6 &79.2  &94.6\\
&MoCo-EMAN (2$\times$) &59.2  &83.7  &69.7  &89.8  &78.9  &94.3\\\cline{2-8}
&MoCo (800) &50.9  &78.1  &66.3  &87.7 &77.2  &93.6\\
&MoCo-EMAN (800) &57.4  &82.3 &68.1  &88.5   &77.4  &93.6\\\cline{2-8}
&BYOL &52.1 &77.3  &67.7 &88.5 &77.0  &93.5\\
&BYOL-EMAN &54.6 &78.6  &68.1 &88.6 &77.1  &93.5\\ [.1em]
\shline
\end{tabular}\vspace{2mm}
\caption{The linear and the finetuning evaluation on ImageNet. The default model is ResNet-50 trained for 200 epochs. ``2$\times$'' means ResNet-50 of 2$\times$ width and ``800'' means 800 epochs.}
\label{tab:linear and finetune}\vspace{-3mm}
\end{table}

\subsection{Self-supervised Evaluation}
\label{subsec:self eval}

We self-supervised pre-train MoCo and BYOL models with EMAN on unlabeled data and then evaluate learned representations on multiple downstream classification tasks.

\paragraph{Linear Classification and Finetuning}

The linear and finetuning evaluation were on different percentages of labeled ImageNet data, including 1\%, 10\% and 100\%. Only the labeled data were used in these experiments. For 1\% (10\%) labels, five (three) different sets of samples were run and the averaged numbers are shown in Table \ref{tab:linear and finetune}. We searched the best learning rate from \{15,30,60\} (\{0.2,0.4,0.8\}) for MoCo (BYOL) linear evaluation, since they are quite sensitive in these experiments. When finetuning, we found it was important to have different learning rates for the pretrained encoder and the randomly initialized top classifier. We thus used learning rate of 1.0 (0.1) for top classifier for 1\% (10\%) labels, and searched the best learning rate from \{0.0001,0.001,0.01\} for the pretrained encoder when finetuning. All experiments were trained for 50 epochs, with learning rate dropped by 10 times at 30th and 40th epoch.

For linear evaluation in Table \ref{tab:linear and finetune}, while EMAN models have comparable performances as the baselines for 100\% labels, they improve over the baselines by 1-2 points of top-1 accuracy for 10\% labels. The gains become bigger (4-5 points) when only 1\% labels are available.
The observations are consistent across different techniques (MoCo and BYOL), different architectures (ResNet-50 and ResNet-50 of 2$\times$ width), and different epochs (200 and 800). Note that the evaluation on 1\%/10\% labels is more practical than that on 100\% labels, since when full dataset is annotated, the advantage of self-supervised pretraining will be reduced. For example, compared with supervised baseline, the self-supervised models are usually worse for 100\% labels, but have significant gains ($>$30/10 points) for 1\%/10\% labels, indicating the self-supervised pretraining is much more useful when there is insufficient supervision available.

Finetuning usually achieved better results than the linear classification, in Table \ref{tab:linear and finetune}, with increasing gains for more annotations, but they could be worse if the hyperparameters are not carefully tuned as introduced above, especially for fewer labels. The gains by EMAN over those strong baselines are still consistent with the linear classification, and even larger in most of the experiments with 1\% labels.

\begin{table}[t]
\tablestyle{1.8pt}{1.2}
\begin{tabular}{lx{30}x{26}x{22}x{22}x{22}x{22}}\shline
\multirow{2}{*}{Method} &\multirow{2}{*}{Arch} &\multirow{2}{*}{Epochs} &\multicolumn{2}{c}{1\% labels} &\multicolumn{2}{c}{10\% labels}\\
& & & top-1 & top-5 & top-1 &top-5\\\hline
Supervised \cite{DBLP:conf/iccv/BeyerZOK19} &res50 &100 &25.4 &48.4  &56.4 &80.4\\ [.1em]
\shline
InstDisc \cite{DBLP:conf/cvpr/WuXYL18} &res50 &-  &- &39.2  &- &77.4\\
PIRL \cite{DBLP:conf/cvpr/MisraM20} &res50 &800  &- &57.2  &- &83.5\\
CPC v2 \cite{henaff2019data} &res161 &-  &52.7 &77.9  &73.1 &91.2\\
MoCo-v2 \cite{chen2020improved} &res50 &800  &50.9  &78.1  &66.3  &87.7 \\
SimCLR \cite{chen2020simple} &res50 &1000 &48.3 &75.5  &65.6 &87.8\\
PCL \cite{li2020prototypical} &res50 &200 &- &75.6  &- &86.2\\
SwAV \cite{caron2020unsupervised} &res50 &800 &53.9 &78.5  &\bf{70.2} &\bf{89.9}\\
BYOL \cite{grill2020bootstrap} &res50 &1000 &53.2 &78.4  &68.8 &89.0\\\hline
MoCo-EMAN &res50 &800 &\bf{57.4} &\bf{82.3}  &68.1 &88.5\\
BYOL-EMAN &res50 &200 &55.1 &78.9  &68.1 &88.6\\ [.1em]\shline
\end{tabular}\vspace{2mm}
\caption{Comparison with other self-supervised models.}
\label{tab:self-supervised}\vspace{-3mm}
\end{table}

\paragraph{Comparison with the State-of-the-art}

The EMAN models were compared with the state-of-the-art self-supervised learning methods for 1\%/10\% labels in Table \ref{tab:self-supervised}.
To have fair comparison, only the results of ResNet-50 was shown where possible. The reported BYOL \cite{grill2020bootstrap} was pretrained for 1000 epochs, with 53.2/68.8 top-1 accuracy for 1\%/10\% labels, but our BYOL-EMAN achieved 55.1/68.1, which was pretrained only for 200 epochs. Our MoCo-EMAN achieved the accuracy of 57.4 for 1\% labels, which is much higher than the other methods in the table, and 68.1 for 10\% labels. Note that, the comparison between these methods is not completely fair. For example, the SwAV \cite{caron2020unsupervised}, with higher accuracy for 10\% labels, used much more expensive multi-crop strategy, which could also benefit our EMAN models.

\begin{table}[t]
\tablestyle{1.8pt}{1.2}
\begin{tabular}{lx{32}|x{32}|x{32}x{32}}\shline
\multirow{2}{*}{Method} &\multirow{2}{*}{Epochs} &kNN &\multicolumn{2}{c}{retrieval}\\
&& top-1 & mAP & recall\\\hline
Supervised &90 &74.8 &57.9 &37.0 \\ [.1em]\shline
MoCo &200 &54.5 &32.4 &18.5\\
MoCo-EMAN &200 &58.0 &39.8 &24.3\\\hline
MoCo &800 &60.0 &41.4 &25.6\\
MoCo-EMAN &800 &62.8 &47.9 &30.5\\\hline
BYOL &200 &62.8 &37.5 &20.1\\
BYOL-EMAN &200 &64.9 &39.8 &20.4\\ [.1em]\shline
InstDisc \cite{DBLP:conf/cvpr/WuXYL18} &- &46.5 &- &-\\
LA \cite{DBLP:conf/iccv/ZhuangZY19} &- &49.4 &- &-\\
PCL \cite{li2020prototypical} &200 &54.5 &39.5$^\dag$ &24.2$^\dag$\\
SwAV \cite{caron2020unsupervised} &800 &59.2 &35.9$^\dag$ &17.5$^\dag$\\[.1em]\shline
\end{tabular}\vspace{2mm}
\caption{The kNN and image retrieval evaluation on ImageNet. $^\dag$ indicates numbers run by us from the pretrained model.}
\label{tab:knn and retrieval}\vspace{-3mm}
\end{table}

\paragraph{kNN Classification and Image Retrieval}

Although the linear classification is a common strategy to evaluate the self-supervised models in recent years \cite{DBLP:conf/cvpr/He0WXG20,chen2020simple,grill2020bootstrap}, it requires additional training, which is not the most direct way to evaluate the representations. Instead, we also compared the kNN accuracies on full \texttt{train}/\texttt{val} data in Table \ref{tab:knn and retrieval}, following \cite{DBLP:conf/cvpr/WuXYL18,DBLP:conf/iccv/ZhuangZY19,li2020prototypical,caron2020unsupervised}. With this more direct evaluation, the EMAN still has consistent improvements over the MoCo and BYOL baselines. And they also outperform  \cite{DBLP:conf/cvpr/WuXYL18,DBLP:conf/iccv/ZhuangZY19} and recent PCL \cite{li2020prototypical} and SwAV \cite{caron2020unsupervised}.

We also evaluate on the task of image retrieval (find the most relevant entries for each query) on ImageNet which also requires no additional training. This task is a practical application of self-supervised pretraining, since the accurate annotations are usually unavailable in many scenarios of image retrieval. We used \texttt{train} as the retrieval database and \texttt{val} as queries, and followed \cite{DBLP:conf/cvpr/YuanW0TJLF20} to use the top 1000 retrievals for the evaluation of mean averaged precision (mAP) and recall. Table \ref{tab:knn and retrieval} shows the EMAN also has consistent and nontrivial improvements over the baselines for this task. The PCL \cite{li2020prototypical} and SwAV \cite{caron2020unsupervised} are compared, but they have shown much worse results.

It has also been shown that the unsupervised learning is still lagging behind supervised learning for kNN classification and image retrieval, although SwAV \cite{caron2020unsupervised} has presented minor gap to supervised learning for linear evaluation. However, the EMAN models can learn better feature representations for these two tasks.

\begin{table}[t]
\tablestyle{1.8pt}{1.2}
\begin{tabular}{llx{30}|x{22}x{22}x{22}x{22}x{22}}\shline
\multicolumn{2}{l}{Method} &Epochs &$k$=1 & $k$=2 & $k$=4 &$k$=8 &$k$=16\\[.1em]\shline
\parbox[t]{2mm}{\multirow{9}{*}{\rotatebox[origin=c]{90}{ImageNet}}} &Supervised &90 &46.8 &57.2 &64.4 &68.6 &71.0 \\\cline{2-8}
&PCL \cite{li2020prototypical}$^\dag$ &200 &29.5 &36.3  &42.3 &46.9 &50.9\\
&SwAV \cite{caron2020unsupervised}$^\dag$ &800 &23.5 &33.6  &43.5 &51.7 &57.8\\\cline{2-8}
&MoCo &200 &22.8 &28.7  &34.7 &40.7 &46.0\\
&MoCo-EMAN &200 &29.3 &36.0  &41.6 &46.9 &50.8\\\cline{2-8}
&MoCo &800 &31.4 &38.3  &44.1 &49.5 &53.9\\
&MoCo-EMAN &800 &35.8 &43.7  &49.8 &54.0 &57.2\\\cline{2-8}
&BYOL &200 &25.6 &34.2  &42.5 &49.4 &54.7\\
&BYOL-EMAN &200 &27.4 &36.8  &45.6 &52.6 &57.5\\[.1em]\shline
\parbox[t]{2mm}{\multirow{6}{*}{\rotatebox[origin=c]{90}{VOC07}}} &Supervised &90 &56.0 &69.6 &74.9 &79.9 &82.7 \\
&PCL \cite{li2020prototypical} &200 &47.9 &59.6  &66.2 &74.5 &78.3\\\cline{2-8}
&MoCo &200 &47.0 &58.9  &65.3 &72.5 &76.3\\
&MoCo-EMAN &200 &50.1 &59.7  &67.2 &74.1 &77.9\\\cline{2-8}
&BYOL &200 &42.8 &55.4  &63.2 &72.8 &77.7\\
&BYOL-EMAN &200 &44.6 &56.5  &65.4 &73.9 &78.8\\[.1em]\shline
\parbox[t]{2mm}{\multirow{4}{*}{\rotatebox[origin=c]{90}{iNaturalist}}}
&MoCo &1000 &21.1 &25.4  &31.3 &36.2 &41.8\\
&MoCo-EMAN &1000 &24.0 &28.4  &33.3 &38.0 &41.7\\\cline{2-8}
&BYOL &200 &16.8 &22.3  &29.0 &35.0 &40.4\\
&BYOL-EMAN &200 &18.0 &23.9  &30.5 &36.3 &41.5\\ [.1em]
\shline
\end{tabular}\vspace{2mm}
\caption{The low-shot evaluation. $^\dag$ indicates numbers run by us from the pretrained model.}
\label{tab:fewshot}\vspace{-3mm}
\end{table}


\paragraph{Low-shot Classification}

Given the superior performances of EMAN in the regimes of few annotations in Table \ref{tab:linear and finetune}, low-shot classification was evaluated, with $k$ samples per class. Following \cite{DBLP:conf/iccv/GoyalM0M19,li2020prototypical}, we trained linear SVMs \cite{DBLP:journals/ml/CortesV95} on top of the frozen representations.
We searched the best SVM cost parameter $C \in 2^{[-5,5]}$, and averaged the numbers of 5 different sets of samples.

The results in Table \ref{tab:fewshot} have demonstrated that EMAN still improves the MoCo/BYOL baselines in low-shot cases, as low as $k=1$ sample per class. For example, in the experiments of ImageNet, MoCo-EMAN is about 4-6 points better than MoCo. The gains for BYOL are smaller, but still 1-3 points. Note that, our MoCo-EMAN can achieve 35.8\% top-1 accuracy for 1000-way 1-shot ImageNet, which is 12.3 points higher than SwAV \cite{caron2020unsupervised}. Pascal VOC2007 \cite{DBLP:journals/ijcv/EveringhamGWWZ10} and iNaturalist \cite{DBLP:conf/cvpr/HornASCSSAPB18} have also been tested. Since the domain of VOC is similar to ImageNet, we directly used the frozen ImageNet representations for VOC experiments. However, it is not the case for iNaturalist, where ImageNet representations have poor performances, so we train MoCo/BYOL from scratch on iNaturalist for 1000/200 epochs. The improvements by EMAN are still consistent on both datasets.

\begin{table}[t]
\tablestyle{1.8pt}{1.2}
\begin{tabular}{lx{56}x{26}x{22}x{22}x{22}x{22}}\shline
\multirow{2}{*}{Method} &\multirow{2}{*}{Pretrained} &\multirow{2}{*}{Schd.} &\multicolumn{2}{c}{1\% labels} &\multicolumn{2}{c}{10\% labels}\\
& & & top-1 & top-5 & top-1 &top-5\\[.1em]\shline
baseline &None &1$\times$ &- &-  &67.1 &86.7\\
EMAN &None &1$\times$ &- &-  &69.2 &88.3\\
baseline &MoCo &1$\times$ &51.2 &73.5  &70.2 &89.0\\
EMAN &MoCo &1$\times$ &58.1 &80.4  &72.0 &90.2\\
EMAN &MoCo-EMAN &1$\times$ &60.9 &82.5  &72.6 &90.5\\\hline
baseline &None &3$\times$ &- &-  &71.1 &88.9\\
EMAN  &None &3$\times$ &- &-  &72.8 &90.3\\
EMAN &MoCo &3$\times$ &61.4 &82.1  &73.9 &91.0\\
EMAN &MoCo-EMAN &3$\times$ &63.0 &83.4  &74.0 &90.9\\[.1em]\shline
\end{tabular}\vspace{2mm}
\caption{The FixMatch results on ImageNet.}
\label{tab:fixmatch}\vspace{-3mm}
\end{table}

\subsection{Semi-supervised Evaluation}

The semi-supervised learning experiments of FixMatch are shown in Table \ref{tab:fixmatch}, where ``1$\times$'' means training 50 (100) epochs with learning rate dropped at 30/40th (60/80th) epoch, for 1\% (10\%) labels. For 10\% labels, the top-1 accuracy is improved to 69.2 by EMAN from the baseline of 67.1. No results were reported for 1\% labels since the default hyperparameters do not work very well. The default FixMatch is trained from scratch. However, as seen in Section \ref{subsec:self eval}, the self-supervised pretrained models can be a significant help for semi-supervised scenarios. Therefore, we also trained FixMatch initialized from the self-suprvised pretrained models, with initial learning rate of 0.003. When finetuned from MoCo, the 1$\times$ baseline FixMatch has 3.1 points of improvement and FixMatch-EMAN 2.8 points for 10\% labels. For 1\% labels, the gains by EMAN over the baseline FixMatch become bigger ($\sim$7 points). When finetuned from MoCo-EMAN, additional gains are available, which is consistent with the observations of Section \ref{subsec:self eval}. Finally, we have trained 3$\times$ longer models with cosine learning rate schedule, and the improvements are still consistent.

\begin{table}[t]
\tablestyle{1.8pt}{1.2}
\begin{tabular}{lx{30}x{22}x{22}x{22}x{22}}\shline
\multirow{2}{*}{Method} &\multirow{2}{*}{Arch} &\multicolumn{2}{c}{1\% labels} &\multicolumn{2}{c}{10\% labels}\\
& & top-1 & top-5 & top-1 &top-5\\\hline
Supervised \cite{DBLP:conf/iccv/BeyerZOK19} &res50 &25.4 &48.4  &56.4 &80.4\\ [.1em]
\shline
Pseudo-label \cite{lee2013pseudo,DBLP:conf/iccv/BeyerZOK19} &res50 &- &51.6  &- &82.4\\
S4L Rotation \cite{DBLP:conf/iccv/BeyerZOK19} &res50 &- &53.4  &- &83.8\\
UDA \cite{xie2019unsupervised} &res50 &- &-  &68.8 &88.5\\
FixMatch \cite{sohn2020fixmatch} &res50 &- &-  &71.5 &89.1\\
SimCLR-v2 \cite{chen2020big} &res50 &60.0* &-  &70.5* &-\\\hline
FixMatch-EMAN &res50 &\bf{63.0} &\bf{83.4}  &\bf{74.0} &\bf{90.9}\\ [.1em]
\shline
\end{tabular}\vspace{2mm}
\caption{The comparison with other semi-supervised models. * means rough numerical estimates from the plots since no exact numbers for ResNet-50, self-distilled, were reported in \cite{chen2020big}.}
\label{tab:semi-supervised}\vspace{-3mm}
\end{table}

\paragraph{Comparison with the State-of-the-art}

The FixMatch-EMAN models are compared with the state-of-the-art semi-supervised methods in Table \ref{tab:semi-supervised}. For 10\% labels
the proposed FixMatch-EMAN achieves 74.0 top-1 accuracy, beating out the original FixMatch \cite{sohn2020fixmatch} by 2.5 points.
Note that this is very close to the fully supervised learning accuracy of 76.1 in Table \ref{tab:linear and finetune}. For 1\% labels, the best previously reported results are SimCLR-v2
of roughly 60.0, with knowledge distillation being trained for 300 epochs after self-supervised pretraining and semi-supervised finetuning. Our FixMatch-EMAN achieved 63.0, which is about 3.0 points higher than SimCLR-v2, with simpler pipeline and fewer epochs (150). Finally, we note the specifically designed semi-supervised learning algorithms (in Table \ref{tab:semi-supervised}) outperform self-supervised pre-trainning followed by semi-superivised finetuning (in Table \ref{tab:self-supervised}) for annotation insufficient scenarios.

\begin{table}[t]
\tablestyle{1.8pt}{1.2}
\begin{tabular}{cc|x{32}x{32}x{32}}\shline
student &teacher &FixMatch &MoCo &BYOL\\ [.1em]
\shline
default &default &67.1 &54.4  &55.4\\
BN &BN &58.9 &52.5  &52.0\\
BN &EMAN &69.2 &55.8  &56.2\\\hline
BN &EMAN ($m$=0.9) &69.0 &51.2  &-\\
BN &EMAN ($m$=0.99999) &54.6 &failed  &-\\\hline
BN &teacher PN ($\alpha$=0.9) &61.4 &54.6  &52.4\\
BN &student PN ($\alpha$=0.9) &68.6 &52.4  &\text{failed}\\
BN &teacher PN ($\alpha$=0.999) &60.9 &55.5  &54.8\\
BN &student PN ($\alpha$=0.999) &69.2 &55.7  &56.2\\[.1em] \shline
\end{tabular}\vspace{2mm}
\caption{The ablation experiments. ``PN'' means proxy norm, $m$ EMAN momentum of (\ref{equ:eman update}) and $\alpha$ BN momentum of (\ref{equ:momentum stat update}).}
\label{tab:ablation}\vspace{-3mm}
\end{table}

\subsection{Ablation Studies}

The ablation experiments, with results in Table \ref{tab:ablation}, followed the experimental settings of Table \ref{tab:norm}. MoCo and BYOL were evaluated by linear classification as in Section \ref{subsec:self eval} on 10\% labeled data.

\paragraph{EMAN Momentum} We have tested different EMAN momentums of (\ref{equ:eman update}), but the momentum for parameter update of (\ref{equ:ema update}) is remained the same 0.999 as in \cite{DBLP:conf/nips/TarvainenV17,DBLP:conf/cvpr/He0WXG20}. When $m=0.9$ of EMAN, the statistics are updated much faster than the parameters, and the accuracy drops for MoCo but remains almost the same for FixMatch. When $m=0.99999$, the statistics are updated much slower, and both MoCo and FixMatch have much worse performances. These have shown that the normalization statistics should well aligned with the parameters to ensure stable performance.

\paragraph{Other EMAN-similar Designs} Two other designs,  achieving similar goals of EMAN in Section \ref{sec:eman}, were also evaluated. Both of them use (\ref{equ:eman forward}) in the teacher during training, but the difference is what proxy statistics $\mu'$ and $\sigma'^2$ to use. The first design is to use the collected proxy statistics of the teacher following (\ref{equ:momentum stat update}) up to the previous iteration.
This is similar to run the inference mode (\ref{equ:BN inference}) during training in standard BN, but update the proxy statistics using (\ref{equ:momentum stat update}) on the fly.
The second design is to simply copy the proxy statistics from the student, by setting $m=0$ in (\ref{equ:eman update}). They are denoted as ``teacher PN'' and ``student PN'' in Table \ref{tab:ablation}, respectively. When using the default BN momentum $\alpha=0.9$, both designs usually lead to worse performance than EMAN. By setting $\alpha=0.999$ to have better aligned statistics with the parameters, better results are available, and the ``student PN'' achieved very close performances to EMAN.

\section{Conclusion}

In this paper, we proposed a simple  normalization technique, exponential moving average normalization (EMAN), for EMA-teacher based semi- and self-supervised learning. It resolves the issues of cross-sample dependency and parameter mismatch when using the standard BN in EMA-teacher framework. This simple design improves the state of the art in semi- and self-supervised learning. These improvements are consistent across different techniques, network architectures, training duration, and datasets, showing that EMAN is generally applicable.

\newpage
\appendix

\section{Appendix}

\begin{figure}[h]
\centering
\centerline{\epsfig{figure=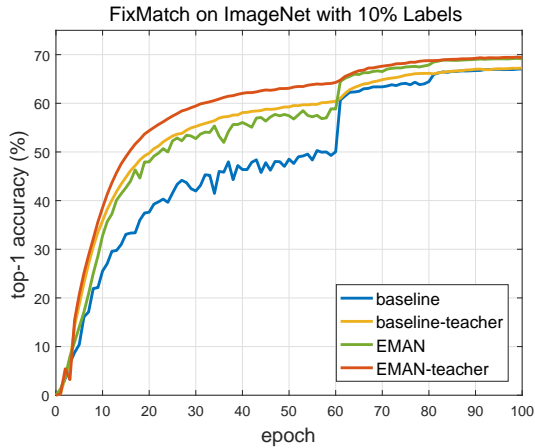,width=8cm,height=6cm}}
\caption{
The FixMatch accuracy with 10\% labels
}
\label{fig:fixmatch_teacher}
\end{figure}

\subsection{FixMatch}

We re-implemented FixMatch in PyTorch, and followed the exactly same hyperparameter settings as in the official FixMatch \cite{sohn2020fixmatch} \footnote{https://github.com/google-research/fixmatch}. The number of labeled (unlabeled) images in a batch is 64 (320). The loss weight for the supervised (unsupervised) loss is 1.0 (10.0). The EMA momentum $m=0.999$. The number of epoch is counted on unlabeled images.

As discussed in \cite{DBLP:conf/iclr/HuangLP0HW17,DBLP:conf/uai/IzmailovPGVW18,DBLP:conf/iclr/AthiwaratkunFIW19}, the EMA updated teacher model can present more reliable results. We show the accuracy curves of the teachers (with EMAN) for both baseline FixMatch and FixMatch-EMAN on ImageNet validation in Figure \ref{fig:fixmatch_teacher}. Although the baseline FixMatch did not use the EMA-teacher framework, we can collect its EMA updated model (both model parameters and BN statistics) during training for the purpose of inference only. It can be found that the EMA teacher indeed has higher and more stable accuracy, especially during the former epochs. This is the reason why we reformulate the baseline FixMatch to EMA-teacher framework.

\begin{figure*}[t]
\begin{minipage}[b]{.33\linewidth}
\centering
\centerline{\epsfig{figure=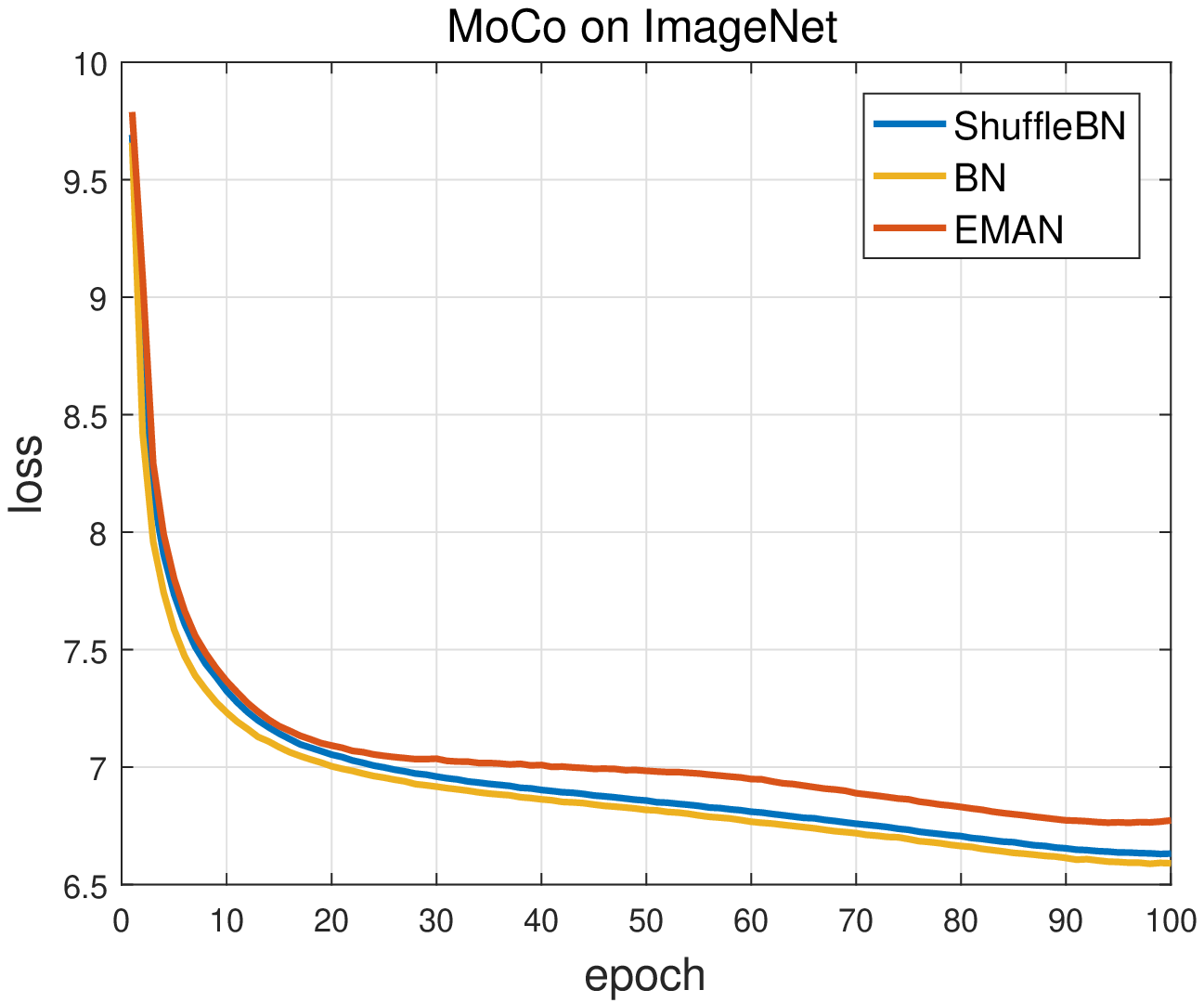,width=6cm,height=4.5cm}}{(a)}
\end{minipage}
\hfill
\begin{minipage}[b]{.33\linewidth}
\centering
\centerline{\epsfig{figure=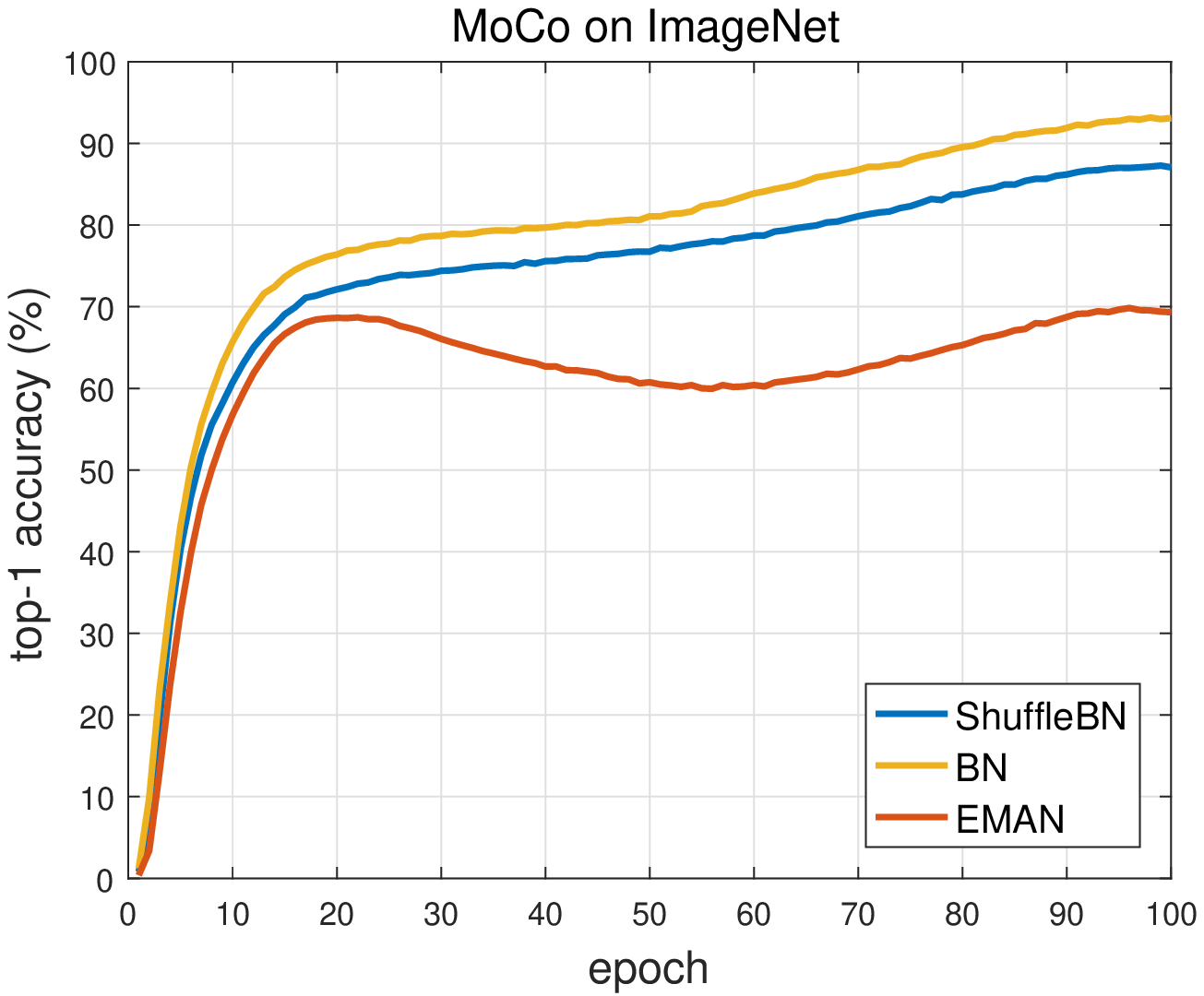,width=6cm,height=4.5cm}}{(b)}
\end{minipage}
\hfill
\begin{minipage}[b]{.33\linewidth}
\centering
\centerline{\epsfig{figure=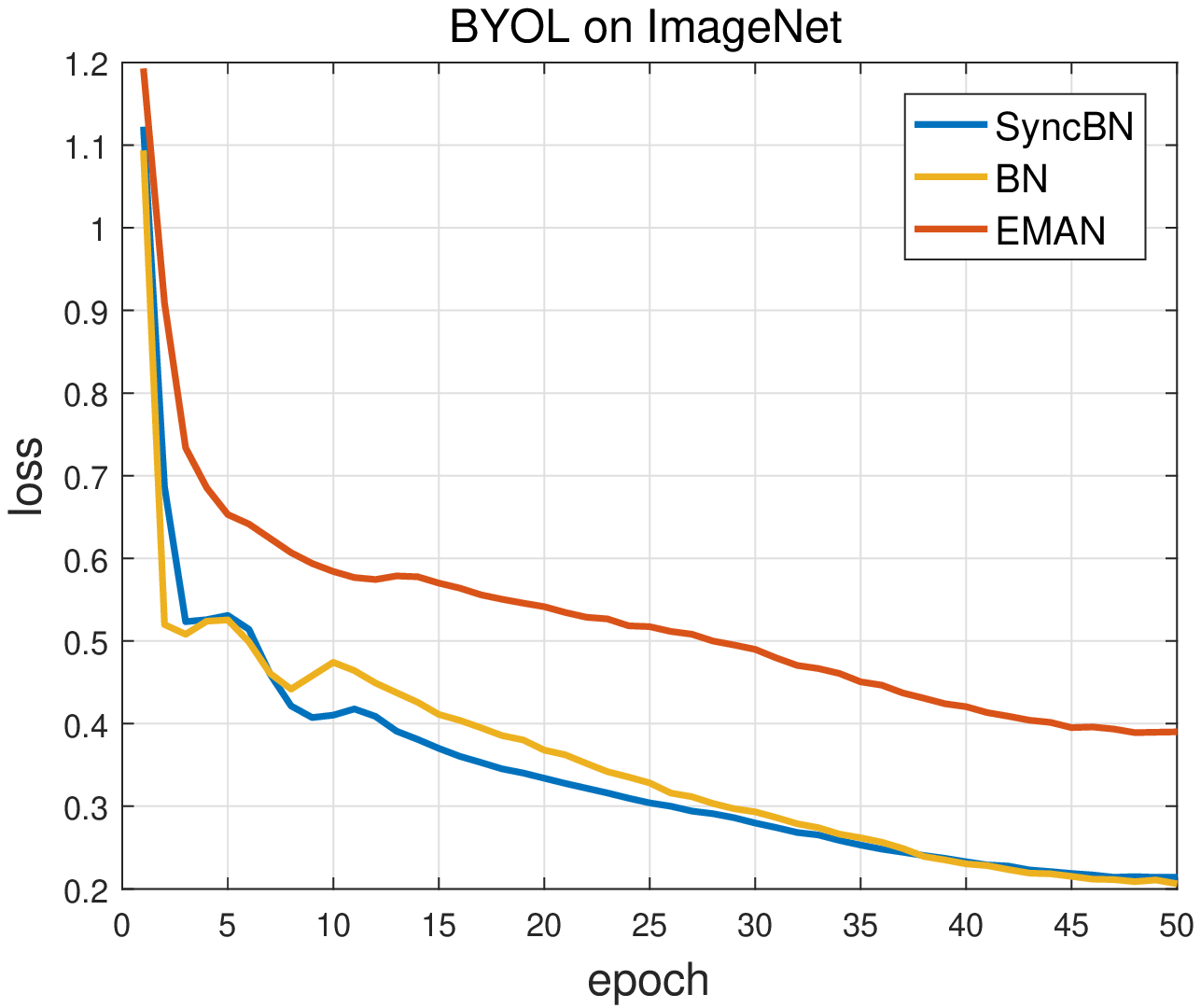,width=6cm,height=4.5cm}}{(c)}
\end{minipage}
\caption{(a) and (b) are the training loss and instance discrimination top-1 accuracy of MoCo, and (c) the training loss of BYOL.
}
\label{fig:loss and acc}
\end{figure*}

\subsection{MoCo}

We used the code and followed the exactly same settings as in the official MoCo-v2 \cite{chen2020improved}\footnote{https://github.com/facebookresearch/moco}. When using EMAN, the training will become less stable at the beginning, because the whole network, including the normalization statistics, is slowly updated, with momentum $m=0.999$. In this case, the warn-up learning rate schedule is more important. Without it, MoCo-EMAN will have worse performance in our experiments. In addition, as seen in Figure \ref{fig:ablation} (b), the kNN accuracy of MoCo-EMAN will have a slight decrease in the last training epochs, because the learning rate is too small. This behavior is quite consistent in our MoCo-EMAN experiments, and the best model is usually at around 90\% training epochs. To avoid the decrease, we simply used the 90\%-th epoch checkpoint for the evaluation of other downstream tasks. Another solution is to set the minimum learning rate to a not-too-small value, e.g. 0.001, in the cosine learning rate schedule. These two strategies achieved very close performances in our experiments.

In Figure \ref{fig:loss and acc} (a) and (b), we show the curves of loss and instance discrimination accuracy of MoCo during training. Although they do not directly relate to the actual representations power, they can help to understand what is happening during training. It can be found that the training behaviors are quite different between MoCo-ShuffleBN and MoCo-EMAN. EMAN will make the self-supervised learning task of MoCo more difficult, and lead to higher training loss and lower instance discrimination accuracy. It suggests that EMAN can better prevent the MoCo model from cheating, and could potentially improve the representation power.

\subsection{BYOL}

We re-implemented BYOL in PyTorch, and followed some hyperparameter settings as in the official BYOL \cite{grill2020bootstrap}\footnote{https://github.com/deepmind/deepmind-research/tree/master/byol}. The official implementation chooses different hyperparameters for experiments of different numbers of epochs. To be consistent, we set weight decay as 0.000001, and initial EMA momentum as 0.98, for experiments of both 50 and 200 epochs. The initial learning rate was set as 0.9 (1.8) for 50 (200) epochs. In all BYOL experiments, the batch size was 512 on a machine with 8 GPUs.

The training loss curves of BYOL are also shown in Figure \ref{fig:loss and acc} (c). Similar to the observations from MoCo curves, EMAN will make the self-supervised learning task of BYOL more difficult, and result in higher training loss. It suggests that EMAN can better prevent the BYOL model from cheating, and could potentially improve the representation power.

\subsection{Other Settings}

\paragraph{ImageNet Sampling} We sample 1\% (10\%) images per class for the semi-supervised learning experiments of 1\% (10\%) labels, which are 12,820 (128,118) images in total.

\paragraph{Linear and Finetuning Evaluation} In addition to the details in Section \ref{subsec:self eval}, weight decay $=0$ and momentum $=0.9$ for linear evaluation, and weight decay $=0.0001$, and momentum $=0.9$ for finetuning. The standard data augmentation (\texttt{RandomResizedCrop} and \texttt{RandomHorizontalFlip}) was used during training. At inference, the center 224$\times$224 crop was used.

\paragraph{kNN Evaluation} Different from \cite{DBLP:conf/cvpr/WuXYL18,DBLP:conf/iccv/ZhuangZY19}, we did not use the weighting mechanism. Instead, it was just a standard kNN classifier with top $k=20$ neighbors, where a query will be classified to the majority class of neighbor samples. The center 224$\times$224 crop was used, also in the experiments of Image Retrieval and Low-shot Classification.

{\small
\bibliographystyle{ieee_fullname}
\bibliography{egbib}

\begin{thebibliography}{10}\itemsep=-1pt

\bibitem{DBLP:conf/iclr/AthiwaratkunFIW19}
Ben Athiwaratkun, Marc Finzi, Pavel Izmailov, and Andrew~Gordon Wilson.
\newblock There are many consistent explanations of unlabeled data: Why you
  should average.
\newblock In {\em ICLR}, 2019.

\bibitem{ba2016layer}
Jimmy~Lei Ba, Jamie~Ryan Kiros, and Geoffrey~E Hinton.
\newblock Layer normalization.
\newblock {\em arXiv preprint arXiv:1607.06450}, 2016.

\bibitem{DBLP:conf/iccv/BeyerZOK19}
Lucas Beyer, Xiaohua Zhai, Avital Oliver, and Alexander Kolesnikov.
\newblock {S4L:} self-supervised semi-supervised learning.
\newblock In {\em ICCV}, pages 1476--1485, 2019.

\bibitem{bucilu2006model}
Cristian Bucilua, Rich Caruana, and Alexandru Niculescu-Mizil.
\newblock Model compression.
\newblock In {\em Proceedings of the 12th ACM SIGKDD international conference
  on Knowledge discovery and data mining}, pages 535--541, 2006.

\bibitem{caron2020unsupervised}
Mathilde Caron, Ishan Misra, Julien Mairal, Priya Goyal, Piotr Bojanowski, and
  Armand Joulin.
\newblock Unsupervised learning of visual features by contrasting cluster
  assignments.
\newblock In {\em NeurIPS}, 2020.

\bibitem{chang2015shapenet}
Angel~X Chang, Thomas Funkhouser, Leonidas Guibas, Pat Hanrahan, Qixing Huang,
  Zimo Li, Silvio Savarese, Manolis Savva, Shuran Song, Hao Su, et~al.
\newblock Shapenet: An information-rich 3d model repository.
\newblock {\em arXiv:1512.03012}, 2015.

\bibitem{chapelle2009semi}
Olivier Chapelle, Bernhard Scholkopf, and Alexander Zien.
\newblock Semi-supervised learning (chapelle, o. et al., eds.; 2006).
\newblock {\em IEEE Transactions on Neural Networks}, 20(3):542--542, 2009.

\bibitem{chen2020simple}
Ting Chen, Simon Kornblith, Mohammad Norouzi, and Geoffrey Hinton.
\newblock A simple framework for contrastive learning of visual
  representations.
\newblock In {\em ICML}, volume 119, pages 1597--1607, 2020.

\bibitem{chen2020big}
Ting Chen, Simon Kornblith, Kevin Swersky, Mohammad Norouzi, and Geoffrey
  Hinton.
\newblock Big self-supervised models are strong semi-supervised learners.
\newblock In {\em NeurIPS}, 2020.

\bibitem{chen2020improved}
Xinlei Chen, Haoqi Fan, Ross Girshick, and Kaiming He.
\newblock Improved baselines with momentum contrastive learning.
\newblock {\em arXiv:2003.04297}, 2020.

\bibitem{DBLP:journals/ml/CortesV95}
Corinna Cortes and Vladimir Vapnik.
\newblock Support-vector networks.
\newblock {\em Mach. Learn.}, 20(3):273--297, 1995.

\bibitem{DBLP:conf/naacl/DevlinCLT19}
Jacob Devlin, Ming{-}Wei Chang, Kenton Lee, and Kristina Toutanova.
\newblock {BERT:} pre-training of deep bidirectional transformers for language
  understanding.
\newblock In {\em NAACL-HLT}, pages 4171--4186, 2019.

\bibitem{DBLP:journals/ijcv/EveringhamGWWZ10}
Mark Everingham, Luc~Van Gool, Christopher K.~I. Williams, John~M. Winn, and
  Andrew Zisserman.
\newblock The pascal visual object classes {(VOC)} challenge.
\newblock {\em Int. J. Comput. Vis.}, 88(2):303--338, 2010.

\bibitem{UnderstandingBYOL}
Abe Fetterman and Josh Albrecht.
\newblock Understanding self-supervised and contrastive learning with bootstrap
  your own latent (byol).
\newblock
  https://untitled-ai.github.io/understanding-self-supervised-contrastive-lear%
ning.html, 2020.

\bibitem{DBLP:conf/iclr/GidarisSK18}
Spyros Gidaris, Praveer Singh, and Nikos Komodakis.
\newblock Unsupervised representation learning by predicting image rotations.
\newblock In {\em ICLR}, 2018.

\bibitem{DBLP:conf/iccv/GoyalM0M19}
Priya Goyal, Dhruv Mahajan, Abhinav Gupta, and Ishan Misra.
\newblock Scaling and benchmarking self-supervised visual representation
  learning.
\newblock In {\em ICCV}, pages 6390--6399, 2019.

\bibitem{grill2020bootstrap}
Jean-Bastien Grill, Florian Strub, Florent Altch{\'e}, Corentin Tallec,
  Pierre~H Richemond, Elena Buchatskaya, Carl Doersch, Bernardo~Avila Pires,
  Zhaohan~Daniel Guo, Mohammad~Gheshlaghi Azar, et~al.
\newblock Bootstrap your own latent: A new approach to self-supervised
  learning.
\newblock In {\em NeurIPS}, 2020.

\bibitem{DBLP:conf/cvpr/HadsellCL06}
Raia Hadsell, Sumit Chopra, and Yann LeCun.
\newblock Dimensionality reduction by learning an invariant mapping.
\newblock In {\em CVPR}, pages 1735--1742, 2006.

\bibitem{DBLP:conf/cvpr/He0WXG20}
Kaiming He, Haoqi Fan, Yuxin Wu, Saining Xie, and Ross~B. Girshick.
\newblock Momentum contrast for unsupervised visual representation learning.
\newblock In {\em CVPR}, pages 9726--9735, 2020.

\bibitem{DBLP:conf/cvpr/HeZRS16}
Kaiming He, Xiangyu Zhang, Shaoqing Ren, and Jian Sun.
\newblock Deep residual learning for image recognition.
\newblock In {\em CVPR}, pages 770--778, 2016.

\bibitem{henaff2019data}
Olivier~J H{\'e}naff, Aravind Srinivas, Jeffrey De~Fauw, Ali Razavi, Carl
  Doersch, SM Eslami, and Aaron van~den Oord.
\newblock Data-efficient image recognition with contrastive predictive coding.
\newblock In {\em ICML}, volume 119, pages 4182--4192, 2020.

\bibitem{DBLP:journals/corr/HintonVD15}
Geoffrey~E. Hinton, Oriol Vinyals, and Jeffrey Dean.
\newblock Distilling the knowledge in a neural network.
\newblock {\em CoRR}, abs/1503.02531, 2015.

\bibitem{DBLP:conf/cvpr/HornASCSSAPB18}
Grant~Van Horn, Oisin~Mac Aodha, Yang Song, Yin Cui, Chen Sun, Alexander
  Shepard, Hartwig Adam, Pietro Perona, and Serge~J. Belongie.
\newblock The inaturalist species classification and detection dataset.
\newblock In {\em CVPR}, pages 8769--8778, 2018.

\bibitem{DBLP:conf/iclr/HuangLP0HW17}
Gao Huang, Yixuan Li, Geoff Pleiss, Zhuang Liu, John~E. Hopcroft, and Kilian~Q.
  Weinberger.
\newblock Snapshot ensembles: Train 1, get {M} for free.
\newblock In {\em ICLR}, 2017.

\bibitem{DBLP:conf/icml/IoffeS15}
Sergey Ioffe and Christian Szegedy.
\newblock Batch normalization: Accelerating deep network training by reducing
  internal covariate shift.
\newblock In {\em ICML}, volume~37, pages 448--456, 2015.

\bibitem{DBLP:conf/uai/IzmailovPGVW18}
Pavel Izmailov, Dmitrii Podoprikhin, Timur Garipov, Dmitry~P. Vetrov, and
  Andrew~Gordon Wilson.
\newblock Averaging weights leads to wider optima and better generalization.
\newblock In Amir Globerson and Ricardo Silva, editors, {\em UAI}, pages
  876--885. {AUAI} Press, 2018.

\bibitem{jing2020self}
Longlong Jing and Yingli Tian.
\newblock Self-supervised visual feature learning with deep neural networks: A
  survey.
\newblock {\em IEEE Trans. Pattern Anal. Mach. Intell.}, 2020.

\bibitem{DBLP:conf/iclr/LaineA17}
Samuli Laine and Timo Aila.
\newblock Temporal ensembling for semi-supervised learning.
\newblock In {\em ICLR}, 2017.

\bibitem{lee2013pseudo}
Dong-Hyun Lee.
\newblock Pseudo-label: The simple and efficient semi-supervised learning
  method for deep neural networks.
\newblock In {\em Workshop on challenges in representation learning, ICML},
  volume~3, 2013.

\bibitem{li2020prototypical}
Junnan Li, Pan Zhou, Caiming Xiong, Richard Socher, and Steven~CH Hoi.
\newblock Prototypical contrastive learning of unsupervised representations.
\newblock {\em arXiv preprint arXiv:2005.04966}, 2020.

\bibitem{DBLP:conf/eccv/LinMBHPRDZ14}
Tsung{-}Yi Lin, Michael Maire, Serge~J. Belongie, James Hays, Pietro Perona,
  Deva Ramanan, Piotr Doll{\'{a}}r, and C.~Lawrence Zitnick.
\newblock Microsoft {COCO:} common objects in context.
\newblock In {\em ECCV}, volume 8693, pages 740--755, 2014.

\bibitem{DBLP:conf/cvpr/MisraM20}
Ishan Misra and Laurens van~der Maaten.
\newblock Self-supervised learning of pretext-invariant representations.
\newblock In {\em CVPR}, pages 6706--6716, 2020.

\bibitem{DBLP:journals/pami/MiyatoMKI19}
Takeru Miyato, Shin{-}ichi Maeda, Masanori Koyama, and Shin Ishii.
\newblock Virtual adversarial training: {A} regularization method for
  supervised and semi-supervised learning.
\newblock {\em {IEEE} Trans. Pattern Anal. Mach. Intell.}, 41(8):1979--1993,
  2019.

\bibitem{DBLP:conf/eccv/NorooziF16}
Mehdi Noroozi and Paolo Favaro.
\newblock Unsupervised learning of visual representations by solving jigsaw
  puzzles.
\newblock In {\em ECCV}, volume 9910, pages 69--84, 2016.

\bibitem{DBLP:conf/nips/PaszkeGMLBCKLGA19}
Adam Paszke, Sam Gross, Francisco Massa, Adam Lerer, James Bradbury, Gregory
  Chanan, Trevor Killeen, Zeming Lin, Natalia Gimelshein, Luca Antiga, Alban
  Desmaison, Andreas K{\"{o}}pf, Edward Yang, Zachary DeVito, Martin Raison,
  Alykhan Tejani, Sasank Chilamkurthy, Benoit Steiner, Lu Fang, Junjie Bai, and
  Soumith Chintala.
\newblock Pytorch: An imperative style, high-performance deep learning library.
\newblock In {\em NeurIPS}, pages 8024--8035, 2019.

\bibitem{radford2018improving}
Alec Radford, Karthik Narasimhan, Tim Salimans, and Ilya Sutskever.
\newblock Improving language understanding by generative pre-training, 2018.

\bibitem{DBLP:journals/ijcv/RussakovskyDSKS15}
Olga Russakovsky, Jia Deng, Hao Su, Jonathan Krause, Sanjeev Satheesh, Sean Ma,
  Zhiheng Huang, Andrej Karpathy, Aditya Khosla, Michael~S. Bernstein,
  Alexander~C. Berg, and Fei{-}Fei Li.
\newblock Imagenet large scale visual recognition challenge.
\newblock {\em Int. J. Comput. Vis.}, 115(3):211--252, 2015.

\bibitem{DBLP:conf/aaai/ShenHX19}
Zhiqiang Shen, Zhankui He, and Xiangyang Xue.
\newblock {MEAL:} multi-model ensemble via adversarial learning.
\newblock In {\em AAAI}, pages 4886--4893. {AAAI} Press, 2019.

\bibitem{sohn2020fixmatch}
Kihyuk Sohn, David Berthelot, Chun-Liang Li, Zizhao Zhang, Nicholas Carlini,
  Ekin~D Cubuk, Alex Kurakin, Han Zhang, and Colin Raffel.
\newblock Fixmatch: Simplifying semi-supervised learning with consistency and
  confidence.
\newblock In {\em NeurIPS}, 2020.

\bibitem{DBLP:journals/jmlr/SrivastavaHKSS14}
Nitish Srivastava, Geoffrey~E. Hinton, Alex Krizhevsky, Ilya Sutskever, and
  Ruslan Salakhutdinov.
\newblock Dropout: a simple way to prevent neural networks from overfitting.
\newblock {\em J. Mach. Learn. Res.}, 15(1):1929--1958, 2014.

\bibitem{DBLP:conf/nips/TarvainenV17}
Antti Tarvainen and Harri Valpola.
\newblock Mean teachers are better role models: Weight-averaged consistency
  targets improve semi-supervised deep learning results.
\newblock In {\em NeurIPS}, pages 1195--1204, 2017.

\bibitem{tian2019contrastive}
Yonglong Tian, Dilip Krishnan, and Phillip Isola.
\newblock Contrastive multiview coding.
\newblock In {\em ECCV}, pages 776--794, 2020.

\bibitem{ulyanov2016instance}
Dmitry Ulyanov, Andrea Vedaldi, and Victor Lempitsky.
\newblock Instance normalization: The missing ingredient for fast stylization.
\newblock {\em arXiv:1607.08022}, 2016.

\bibitem{DBLP:conf/eccv/WuH18}
Yuxin Wu and Kaiming He.
\newblock Group normalization.
\newblock In {\em ECCV}, volume 11217, pages 3--19, 2018.

\bibitem{DBLP:conf/cvpr/WuXYL18}
Zhirong Wu, Yuanjun Xiong, Stella~X. Yu, and Dahua Lin.
\newblock Unsupervised feature learning via non-parametric instance
  discrimination.
\newblock In {\em CVPR}, pages 3733--3742, 2018.

\bibitem{xie2019unsupervised}
Qizhe Xie, Zihang Dai, Eduard Hovy, Minh-Thang Luong, and Quoc~V Le.
\newblock Unsupervised data augmentation for consistency training.
\newblock In {\em NeurIPS}, 2020.

\bibitem{DBLP:conf/cvpr/XieLHL20}
Qizhe Xie, Minh{-}Thang Luong, Eduard~H. Hovy, and Quoc~V. Le.
\newblock Self-training with noisy student improves imagenet classification.
\newblock In {\em CVPR}, pages 10684--10695. {IEEE}, 2020.

\bibitem{Yan2020Towards}
Junjie Yan, Ruosi Wan, Xiangyu Zhang, Wei Zhang, Yichen Wei, and Jian Sun.
\newblock Towards stabilizing batch statistics in backward propagation of batch
  normalization.
\newblock In {\em ICLR}, 2020.

\bibitem{DBLP:conf/cvpr/YuanW0TJLF20}
Li Yuan, Tao Wang, Xiaopeng Zhang, Francis E.~H. Tay, Zequn Jie, Wei Liu, and
  Jiashi Feng.
\newblock Central similarity quantization for efficient image and video
  retrieval.
\newblock In {\em CVPR}, pages 3080--3089, 2020.

\bibitem{DBLP:conf/eccv/ZhangIE16}
Richard Zhang, Phillip Isola, and Alexei~A. Efros.
\newblock Colorful image colorization.
\newblock In {\em ECCV}, volume 9907, pages 649--666, 2016.

\bibitem{zhu2005semi}
Xiaojin~Jerry Zhu.
\newblock Semi-supervised learning literature survey.
\newblock Technical report, University of Wisconsin-Madison Department of
  Computer Sciences, 2005.

\bibitem{DBLP:conf/iccv/ZhuangZY19}
Chengxu Zhuang, Alex~Lin Zhai, and Daniel Yamins.
\newblock Local aggregation for unsupervised learning of visual embeddings.
\newblock In {\em ICCV}, pages 6001--6011. {IEEE}, 2019.

\end{thebibliography}
}

\end{document}